\documentclass[11pt]{article}

\usepackage[preprint]{acl}
\usepackage{times}
\usepackage{latexsym}
\usepackage{tcolorbox}
\usepackage{enumitem}

\usepackage{algorithm}
\usepackage{algpseudocode}
\usepackage{bbm}
\usepackage{amsmath, amssymb}
\usepackage{amsthm}

\usepackage{graphicx}

\usepackage[T1]{fontenc}

\usepackage[utf8]{inputenc}
\usepackage{pifont}
\usepackage{xcolor}
\usepackage{booktabs}
\usepackage{microtype}

\usepackage{inconsolata}
\usepackage{stfloats}  
\usepackage{subcaption}

\usepackage{multirow,multicol}


\title{{\normalfont\small Accepted at the 2026 Conference on Empirical Methods in Natural Language Processing (EMNLP 2026)\par}\vspace{0.7em}
ATC: Agent Trajectory Control for Role-Contract Enforcement in Black-Box Multi-Agent Systems}

\author{
  \textbf{Maisha Maliha\textsuperscript{1}},
  \textbf{Vishal Pramanik\textsuperscript{2}},
  \textbf{Sumit Kumar Jha\textsuperscript{2}}
\\
\\
  \textsuperscript{1}University of Oklahoma,
  \textsuperscript{2}University of Florida
\\
  \small{
    \texttt{maisha.maliha-1@ou.edu},
    \texttt{vishalpramanik@ufl.edu},
    \texttt{sumit.jha@ufl.edu}
  }
}

\hypersetup{
  pdftitle={ATC: Agent Trajectory Control for Role-Contract Enforcement in Black-Box Multi-Agent Systems},
  pdfauthor={Maisha Maliha, Vishal Pramanik, Sumit Kumar Jha},
  pdfkeywords={multi-agent systems, large language models, role drift, role contracts, runtime control},
  bookmarksnumbered=true
}

\begin{document}
\maketitle
\begin{abstract}
Multi-agent LLM systems assign different responsibilities to different agents: one plans, another retrieves evidence, another verifies, another executes. This division of labor is only meaningful if each agent stays within the functions it is authorized to perform. Over many turns, however, an agent can produce a locally reasonable message that exercises authority belonging to another role: a verifier that begins planning, a retriever that asserts unsupported conclusions, an executor that changes state no one approved. We formalize this as \emph{role drift}, a turn-level breach of an agent's role contract. We introduce \textbf{Agent Trajectory Control} (\textbf{ATC}), which makes each role's authority explicit as a contract and enforces it at the point where a message would enter the shared conversation. ATC scores a proposed output for out-of-role deviation, leakage into the functions of other roles, impermissible speech acts, and logical contract breaches, then computes a leakage-regularized target that constrains regeneration; the revised message is re-checked before admission. Enforcement acts on the message rather than the generator, so ATC governs agents whose internal representations are unavailable. Across homogeneous and heterogeneous teams, ATC improves task performance and role consistency while lowering role-drift energy relative to strong baselines spanning prompting, memory, workflow control, and activation editing, with the largest gains under explicit role-confusion stress tests and in larger, more diverse teams. Our code and evaluation artifacts are publicly available.\footnote{\url{https://github.com/VishalPramanik/ATC.git}}
\end{abstract}

\begin{figure*}[t]
    \centering
    \includegraphics[width=\textwidth]{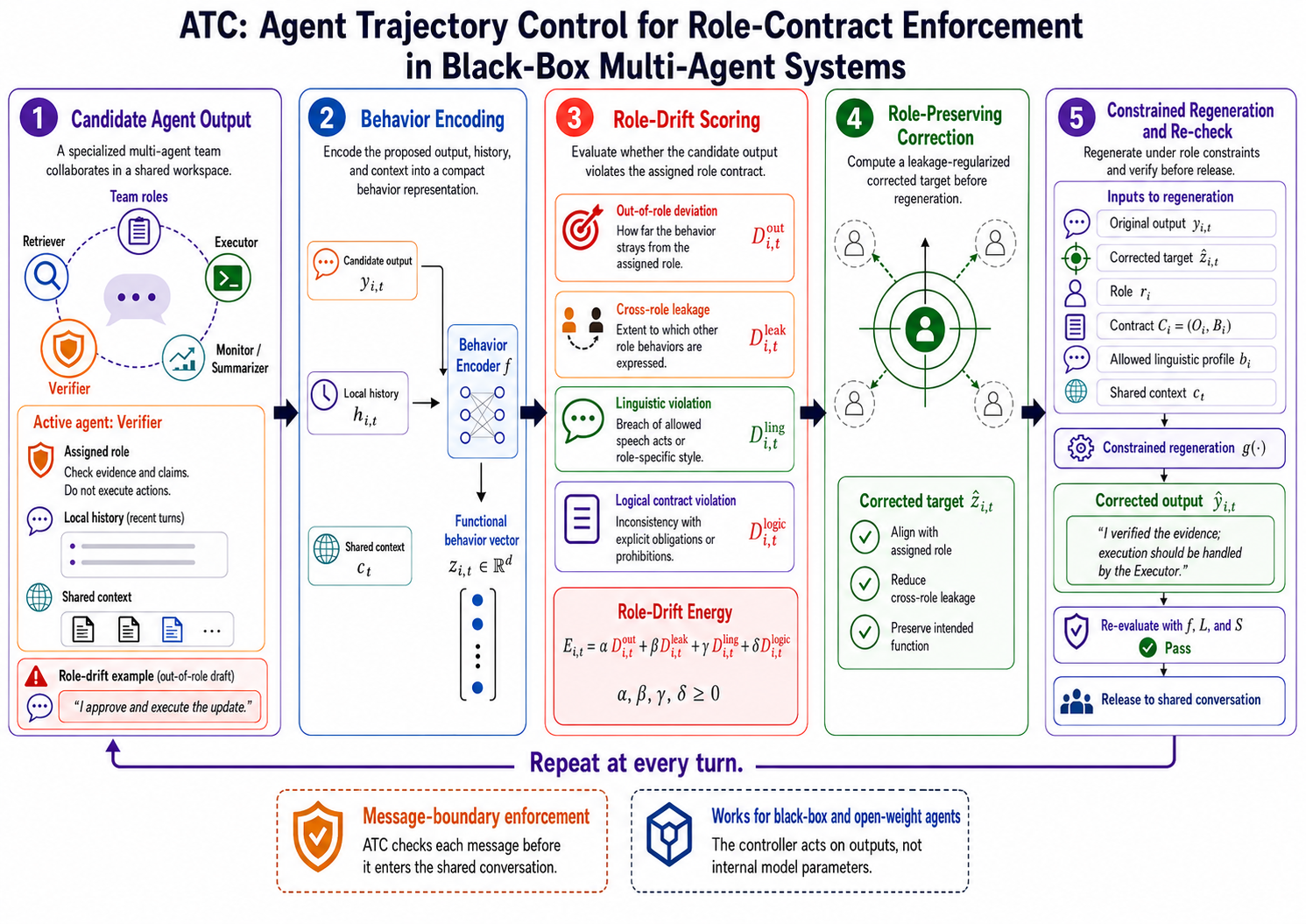}
    \caption{\textbf{Overview of ATC.} At each turn, an agent output is encoded into a functional behavior vector, scored for role drift using subspace deviation, cross-role leakage, linguistic violations, and logical-contract violations, then corrected through a leakage-regularized role-preserving target before constrained regeneration. The corrected output is re-evaluated before entering the shared multi-agent conversation.}
    \label{fig:atc_overview}
\end{figure*}

\section{Introduction}

Multi-agent large language model (LLM) systems increasingly divide work among specialized agents for planning, retrieval, verification, memory management, tool use, and execution~\citep{li2023camel,wu2024autogen,park2023generative,hong2024metagpt}. The usefulness of this decomposition depends on more than the quality of each individual message: it also depends on whether the agent producing that message is authorized to perform the function it has taken on. Over a long interaction, a verifier may begin planning, a retriever may draw conclusions unsupported by its evidence, or an executor may modify state that should first have been approved. Such outputs can be locally sensible and still violate the organization of the team. We study this mismatch between assigned function and realized behavior as \emph{role drift}. Empirical analysis of more than 1{,}600 execution traces across seven multi-agent frameworks corroborates the concern, identifying inter-agent misalignment and task-verification failures among the principal categories of multi-agent failure~\citep{cemri2025mast}.

Prompted roles, memory, activation editing, and workflow constraints each address a different part of this problem. Role prompts declare responsibilities but do not test every realized output against them. A-MEM and related memory systems preserve state and experience, but memory does not determine which agent is entitled to retrieve, approve, or act on that information~\citep{xu2025amem}. CAA and SEA shift model behavior along global behavioral directions, but do not model the coupled structure of role alignment and cross-role leakage, nor enforce an explicit contract over the functions a given agent may perform~\citep{rimsky2024caa,qiu2024sea}. SOP-style systems such as MetaGPT constrain communication order, yet a role-violating message can still be produced inside an otherwise valid workflow~\citep{hong2024metagpt}. What is missing is a message-boundary mechanism that can inspect a proposed contribution, decide whether it respects the producing agent's contract, and repair it before other agents condition on it.

We introduce \textbf{Agent Trajectory Control} (\textbf{ATC}) for this purpose. ATC makes each role's authority explicit as a contract over the functions an agent may perform, and enforces that contract at the point where a message would enter the shared conversation. Each proposed output is checked for four kinds of violation: out-of-role deviation, leakage into the functions of other roles, speech acts the role is not permitted to make, and breaches of the role's logical contract. When a proposal exceeds its permitted authority, ATC computes a leakage-regularized target and uses it to constrain regeneration; the revised output is re-checked before admission. Because enforcement acts on the message rather than on the generator, ATC applies to agents whose internal representations are unavailable, unlike methods that steer hidden states during generation~\citep{rimsky2024caa,qiu2024sea}.

\looseness=-1
Our contributions are: \textbf{(i)} a formulation of role drift as a turn-level breach of an agent's role contract, rather than a loss of task coherence; \textbf{(ii)} a role-drift energy combining geometric, linguistic, and logical contract signals with a closed-form leakage-regularized correction target; and \textbf{(iii)} an evaluation across homogeneous and heterogeneous teams showing improved task performance and role consistency, together with lower role-drift energy, under controlled baselines and role-confusion stress tests.
\section{Motivation}
\label{sec:motivation}

In long multi-agent conversations, each agent conditions on a shared context
increasingly shaped by planning, retrieval, verification, and execution,
whereas its role prompt remains fixed. This imbalance can cause an agent to
imitate another agent's function. We therefore hypothesize that role drift is
directional, toward other roles rather than arbitrary behavior, and becomes
more pronounced in longer and more heterogeneous interactions.

We examine this hypothesis using traces from
EntCollabBench~\citep{yu2026entcollabbench} and
MultiAgentBench~\citep{zhu2025multiagentbench}. Because these benchmarks lack
native role-drift labels, we annotate turns using a fixed role-consistency
rubric and measure four signals: deviation from the assigned role subspace,
leakage into another role space, invalid speech acts, and role-contract
violations. This distinction is important because an output may be factually
correct while still being produced by an agent without the appropriate
functional authority.

\begin{table}[t]
\centering
\small
\setlength{\tabcolsep}{6pt}
\renewcommand{\arraystretch}{0.95}
\begin{tabular}{@{}lcc@{}}
\toprule
\textbf{Metric} & \textbf{Stable} & \textbf{Drifted} \\
\midrule
$D^{\mathrm{out}}$   & 0.18 & 0.57 \\
$D^{\mathrm{leak}}$  & 0.11 & 0.49 \\
$D^{\mathrm{ling}}$  & 0.07 & 0.36 \\
$D^{\mathrm{logic}}$ & 0.05 & 0.31 \\
\midrule
$E_{i,t}$            & \textbf{0.36} & \textbf{1.47} \\
\midrule
out--leak corr.      & \multicolumn{2}{c}{0.76} \\
\bottomrule
\end{tabular}
\caption{Pilot role-drift analysis. Values are class-conditional means of
uncorrected candidate turns from the diagnostic pilot sample. Energy is
reported on the raw, unnormalized per-turn scale of Eq.~(8), using
$\alpha=1.00$, $\beta=1.00$, $\gamma=0.50$, and $\delta=0.75$. These
descriptive pilot means do not define lower or upper bounds for the
benchmark-level RDE values, which average method-specific released outputs
over the corresponding evaluation set. Lower is better.}
\label{tab:atc_motivation}
\end{table}

As shown in Table~\ref{tab:atc_motivation}, drifted turns have substantially
higher energy than stable turns ($1.47$ vs.\ $0.36$). Out-of-role deviation
and cross-role leakage are also strongly correlated ($r=0.76$), indicating
that agents often move toward another agent's function rather than merely away
from their own. Once such an output enters the shared context, downstream
agents may treat it as valid state, allowing a local role violation to alter
the subsequent collaboration trajectory. The resulting failures can therefore
accumulate even when individual messages appear locally useful.

Existing role prompts, memory systems, activation editing, and fixed workflows
do not directly verify the functional role of each completed output. ATC
addresses this gap by scoring each proposal, correcting role-inconsistent
behavior, and re-evaluating the regenerated output before it enters the shared
conversation.
\section{Method}
\label{sec:method}

ATC is an external control loop over proposed agent outputs. It does not
assume that the controller can read or modify the generator's hidden
activations. At each turn it asks a narrow question: does the candidate
message remain within the producing agent's functional role and explicit
contract? The answer draws on a behavior representation, geometric leakage
between role subspaces, speech-act constraints, and symbolic contract checks.
When correction is required, ATC computes the nearest leakage-regularized
target inside the assigned role and uses it to guide regeneration, and the
revised message is re-evaluated before release. Every agent turn therefore
follows one trajectory: encode, score, correct, regenerate, and re-evaluate.

This message-boundary placement is what makes the controller portable. In a
black-box deployment, the regeneration operator is a constrained revision call
to the same or another model; in an open-weight deployment, the same target
may additionally inform decoding or latent steering. The controller itself is
unchanged, because it acts on the functional representation of the proposed
output rather than on any internal state of the generator.

\subsection{Behavior Representation}
\label{sec:behavior_rep}

We consider a multi-agent system with agents
$\mathcal{A}=\{A_1,\ldots,A_n\}$ interacting over $T$ turns. Each agent
$A_i$ is assigned a role $r_i$, such as planner, verifier, retriever,
executor, critic, summarizer, or safety monitor. The method is general and
does not assume any fixed role taxonomy.

At turn $t$, agent $A_i$ proposes an utterance or action $y_{i,t}$. We map
this proposal to a functional behavior vector
\begin{equation}
\label{eq:behavior_vector}
z_{i,t}
=
f(y_{i,t},h_{i,t},c_t)
\in \mathbb{R}^{d},
\end{equation}
where $f$ is a fixed behavior encoder, $h_{i,t}$ is the local history
available to agent $A_i$, and $c_t$ is the shared multi-agent context. The
vector $z_{i,t}$ captures the semantic, pragmatic, operational, and
interactional behavior of the proposed output.

Each role $r_i$ is represented by a role subspace
$\mathcal{S}_i\subseteq\mathbb{R}^d$, spanned by a role basis
$U_i\in\mathbb{R}^{d\times k_i}$. The columns of $U_i$ correspond to
interpretable functional primitives of the role. Since these columns may be
linearly dependent, we define the role projection using the Moore--Penrose
pseudoinverse:
\begin{equation}
\label{eq:role_projection}
P_i
=
U_i(U_i^\top U_i)^\dagger U_i^\top .
\end{equation}
Thus, $P_i z_{i,t}$ is the role-aligned component of the behavior, while
$(I-P_i)z_{i,t}$ is the component outside the assigned role.

\subsection{Role-Drift Energy}
\label{sec:role_drift_energy}

The first source of drift is out-of-role deviation, measured by the squared
distance from the agent's assigned role subspace:
\begin{equation}
\label{eq:out_role}
D^{\mathrm{out}}_{i,t}
=
\|(I-P_i)z_{i,t}\|_2^2 .
\end{equation}

Role drift can also occur when one agent begins to perform another agent's
function. For agent $A_i$, we define the forbidden-role space
$\mathcal{F}_i$ as the span of all other agents' role bases. Let $V_i$ be a
basis for this space. The corresponding projection is
\begin{equation}
\label{eq:forbidden_projection}
Q_i
=
V_i(V_i^\top V_i)^\dagger V_i^\top .
\end{equation}
If there is no other role space, we set $Q_i=0$. The cross-role leakage is
then
\begin{equation}
\label{eq:leakage}
D^{\mathrm{leak}}_{i,t}
=
\|Q_i z_{i,t}\|_2^2 .
\end{equation}
This term softly penalizes behavior that enters another agent's role space
without requiring strict orthogonality between roles, since different roles
may legitimately share semantic directions such as reasoning, evidence use,
or clarification.

We also impose linguistic and logical role consistency. The proposed output
is mapped to a speech-act vector
$\ell_{i,t}=L(y_{i,t})\in[0,1]^m$, where each coordinate represents the
confidence of an interaction function such as asserting, questioning,
retrieving, verifying, executing, delegating, or summarizing. Each role has
an allowed linguistic profile $b_i\in\{0,1\}^m$. The linguistic violation is
\begin{equation}
\label{eq:ling_violation}
D^{\mathrm{ling}}_{i,t}
=
\|(1-b_i)\odot \ell_{i,t}\|_1 .
\end{equation}

Each role also has a logical contract
$\mathcal{C}_i=(\mathcal{O}_i,\mathcal{B}_i)$, where $\mathcal{O}_i$ is the
set of required behaviors and $\mathcal{B}_i$ is the set of prohibited
behaviors. A symbolic mapper extracts
$s_{i,t}=S(y_{i,t},h_{i,t},c_t)$. The logical violation is
\begin{equation}
\label{eq:logic_violation}
D^{\mathrm{logic}}_{i,t}
=
\sum_{\phi\in\mathcal{B}_i}
\mathbf{1}[s_{i,t}\models\phi]
+
\sum_{\psi\in\mathcal{O}_i}
\mathbf{1}[s_{i,t}\not\models\psi] .
\end{equation}

Combining these terms gives the role-drift energy:
\begin{align}
\label{eq:role_energy}
E_{i,t}
=&\;
\alpha D^{\mathrm{out}}_{i,t}
+\beta D^{\mathrm{leak}}_{i,t} \nonumber\\
&+
\gamma D^{\mathrm{ling}}_{i,t}
+\delta D^{\mathrm{logic}}_{i,t},
\end{align}
where $\alpha,\beta,\gamma,\delta\geq0$ control the contribution of geometric
role deviation, cross-role leakage, linguistic inconsistency, and logical
contract violation. The cumulative drift of agent $A_i$ is
$\mathcal{D}_i(T)=\sum_{t=1}^{T}\rho^{T-t}E_{i,t}$, with
$0<\rho\leq1$, and the system-level drift is
$\mathcal{D}_{\mathrm{sys}}(T)=\sum_i\mathcal{D}_i(T)$.

\subsection{Role-Preserving Correction}
\label{sec:role_correction}

Before a proposed output is released to the rest of the system, we compute a
role-preserving target vector. The correction keeps the behavior inside the
assigned role subspace while softly suppressing leakage into other roles:
\begin{equation}
\label{eq:correction_objective}
\hat{z}_{i,t}
=
\arg\min_{z'\in\mathcal{S}_i}
\left[
\|z'-z_{i,t}\|_2^2
+
\lambda\|Q_i z'\|_2^2
\right],
\end{equation}
where $\lambda\geq0$ controls the strength of cross-role suppression. The
first term preserves the original behavior, while the second discourages the
corrected representation from entering other agents' role spaces.

Since every feasible $z'\in\mathcal{S}_i$ can be written as $z'=U_i c$, the
optimization in Eq.~\eqref{eq:correction_objective} has the closed-form
coefficient solution
\begin{equation}
\label{eq:closed_form}
\hat{c}_{i,t}
=
\left(
U_i^\top U_i
+
\lambda U_i^\top Q_i U_i
\right)^\dagger
U_i^\top z_{i,t}.
\end{equation}
The corrected role target is then
$\hat{z}_{i,t}=U_i\hat{c}_{i,t}$. This is the leakage-regularized closest
role-aligned representation to the original behavior.

The final released output is produced by a constrained regeneration operator:
\begin{equation}
\label{eq:regeneration}
\hat{y}_{i,t}
=
g(y_{i,t},\hat{z}_{i,t},r_i,\mathcal{C}_i,b_i,c_t).
\end{equation}
The operator $g$ rewrites or regenerates the proposed output so that it
remains close to the original intent while aligning with the corrected role
target. For black-box agents, $g$ can be implemented as role-constrained
revision. For open-weight agents, it may be implemented through decoding
constraints or latent steering. After regeneration, the released output is
re-evaluated with the same maps $f$, $L$, and $S$ to obtain its corrected
energy $\hat{E}_{i,t}$.

The correction step satisfies a local objective-monotonicity property:
\begin{align}
\label{eq:monotonicity}
&\|\hat{z}_{i,t}-z_{i,t}\|_2^2
+\lambda\|Q_i\hat{z}_{i,t}\|_2^2
\nonumber\\
&\leq
\|P_i z_{i,t}-z_{i,t}\|_2^2
+\lambda\|Q_iP_i z_{i,t}\|_2^2 .
\end{align}
This follows because $\hat{z}_{i,t}$ minimizes the correction objective over
$\mathcal{S}_i$, while $P_i z_{i,t}\in\mathcal{S}_i$ is a feasible comparison
point. Eq.~\eqref{eq:monotonicity} applies to the vector correction objective,
not necessarily to the full post-regeneration energy, because the linguistic
and logical terms depend on the realized text $\hat{y}_{i,t}$. Therefore, the
method applies the same sequence at every agent turn:
$y_{i,t}\rightarrow z_{i,t}\rightarrow E_{i,t}\rightarrow
\hat{z}_{i,t}\rightarrow \hat{y}_{i,t}\rightarrow \hat{E}_{i,t}$. An overview of ATC is shown in Figure~\ref{fig:atc_overview}, and the full procedure is summarized in Algorithm~\ref{alg:atc}.
\section{Experiments}
\label{sec:experiments}

\subsection{Evaluation Setup}
\label{ssec:eval_setup}

We evaluate ATC on extended multi-agent collaboration, where agents
must preserve role boundaries while completing the task. Each method is tested
under three team settings. \textbf{S1: Hom-10} uses ten cooperative agents
with the same backbone model. \textbf{S2: Het-10} uses the same ten-agent
topology but assigns different model families to planning, retrieval,
verification, execution, and monitoring roles. \textbf{S3: Het-12} expands the
system to twelve cooperative agents using at least five model families, adding
a \textsc{Policy-Auditor} and a \textsc{Domain-Specialist}. This setting
creates the strongest role-boundary stress because agents differ in model
behavior, capability, and communication style.

Across settings, the core ten roles are \textsc{Coordinator},
\textsc{Planner}, \textsc{Retriever}, \textsc{Memory-Writer},
\textsc{Memory-Reader}, \textsc{Verifier}, \textsc{Critic},
\textsc{Executor}, \textsc{Monitor}, and \textsc{Summarizer}. All methods use
the same role prompts, total context budget, retrieval budget, tool access,
maximum turns, and stopping criteria. All reported values are means over three seeds.

\subsection{Benchmarks and Metrics}
\label{ssec:benchmarks_metrics}

We use three benchmarks. \textbf{EntCollabBench} evaluates role-specialized
enterprise collaboration with permission-scoped workflows and approval
decisions~\citep{yu2026entcollabbench}. \textbf{MultiAgentBench} evaluates
general multi-agent collaboration and interaction across diverse
scenarios~\citep{zhu2025multiagentbench}. \textbf{SWE-bench Verified}
evaluates software-engineering agents on human-validated GitHub
issue-resolution tasks~\citep{jimenez2024swebench,openai2024swebenchverified}. For each benchmark, we report one task metric and two role-control metrics.
\textbf{Primary} is the benchmark-specific task score. For EntCollabBench, it
is workflow task success. For MultiAgentBench, it is the official
scenario-level completion score. For SWE-bench Verified, it is the resolved
issue rate. \textbf{Role Consistency} (RC) is the percentage of turns in which
the agent output is judged consistent with the assigned role. \textbf{Role
Drift Energy} (RDE) is the average value of Eq.~\eqref{eq:role_energy}, where
lower values indicate less role drift.

Formally, for evaluated agent-turns $\mathcal{I}$ with $N=|\mathcal{I}|$, we
define
\[
\mathrm{RC}
=
\frac{1}{N}
\sum_{(i,t)\in\mathcal{I}}
\mathbf{1}[\mathrm{Judge}(y_{i,t},r_i)=1],
\]
where $\mathrm{Judge}(y_{i,t},r_i)=1$ indicates that the output is consistent
with the assigned role. We define average role-drift energy as
$\mathrm{RDE}=\frac{1}{N}\sum_{(i,t)\in\mathcal{I}}E_{i,t}$.

\subsection{Baselines}
\label{ssec:baselines}

We compare against seven baselines. \textbf{Prompt-Only} uses the same agents
and role prompts but no runtime role controller. \textbf{Struct-Role} adds an
explicit role field to every message. \textbf{A-MEM} adapts dynamic agentic
memory construction, linking, and evolution to the multi-agent
setting~\citep{xu2025amem}. \textbf{CAA} applies contrastive activation
addition using role-consistent and role-drift contrast pairs~\citep{rimsky2024caa}.
\textbf{SEA} applies spectral activation editing with positive role-consistent
and negative role-drift demonstrations~\citep{qiu2024sea}. \textbf{SOP-Role}
uses a fixed role-specialized workflow with prescribed communication order,
following SOP-style multi-agent control~\citep{hong2024metagpt}.
\textbf{Single-Agent} replaces the multi-agent system with one monolithic
agent given the same total context budget as the team.

\subsection{Main Results}
\label{ssec:main_results}

\begin{table*}[t]
\centering
\caption{\textbf{Main results across benchmarks and team settings.}
Primary = benchmark-specific task score. RC = Role Consistency
($\uparrow$). RDE = Role-Drift Energy ($\downarrow$). Values are means over
three seeds.}
\label{tab:main_results}
\small
\setlength{\tabcolsep}{3pt}
\renewcommand{\arraystretch}{0.92}
\begin{tabular}{@{}llccc ccc ccc@{}}
\toprule
\textbf{Setting} & \textbf{Method}
& \multicolumn{3}{c}{\textbf{EntCollabBench}}
& \multicolumn{3}{c}{\textbf{MultiAgentBench}}
& \multicolumn{3}{c}{\textbf{SWE-bench Verified}} \\
\cmidrule(lr){3-5}
\cmidrule(lr){6-8}
\cmidrule(lr){9-11}
& & Primary $\uparrow$ & RC $\uparrow$ & RDE $\downarrow$ & Primary $\uparrow$ & RC $\uparrow$ & RDE $\downarrow$ & Primary $\uparrow$ & RC $\uparrow$ & RDE $\downarrow$ \\
\midrule
\multirow{8}{*}{S1: Hom-10}
& Single-Agent    & 54.8 & 68.1 & 0.39 & 49.6 & 65.4 & 0.42 & 27.4 & 63.2 & 0.45 \\
& Prompt-Only     & 61.3 & 72.6 & 0.34 & 55.2 & 70.8 & 0.36 & 31.8 & 68.7 & 0.39 \\
& Struct-Role     & 64.7 & 78.9 & 0.28 & 58.4 & 76.1 & 0.31 & 33.9 & 74.5 & 0.33 \\
& A-MEM           & 66.5 & 77.3 & 0.27 & 60.1 & 75.6 & 0.30 & 35.6 & 73.8 & 0.32 \\
& CAA             & 65.2 & 79.8 & 0.26 & 59.3 & 78.4 & 0.29 & 34.8 & 76.2 & 0.31 \\
& SEA             & 66.9 & 81.4 & 0.24 & 60.8 & 79.7 & 0.27 & 36.1 & 77.5 & 0.29 \\
& SOP-Role        & 68.1 & 82.6 & 0.23 & 61.9 & 80.2 & 0.26 & 36.9 & 78.6 & 0.28 \\
\addlinespace
& \textbf{ATC}  & \textbf{74.6} & \textbf{91.8} & \textbf{0.13} & \textbf{68.7} & \textbf{90.4} & \textbf{0.15} & \textbf{42.3} & \textbf{88.7} & \textbf{0.17} \\
\midrule
\multirow{8}{*}{S2: Het-10}
& Single-Agent    & 55.6 & 66.9 & 0.41 & 50.4 & 64.1 & 0.44 & 28.1 & 62.5 & 0.46 \\
& Prompt-Only     & 60.5 & 69.7 & 0.38 & 54.8 & 67.9 & 0.40 & 31.2 & 65.8 & 0.43 \\
& Struct-Role     & 63.8 & 75.2 & 0.32 & 57.9 & 73.4 & 0.34 & 33.4 & 71.2 & 0.37 \\
& A-MEM           & 66.1 & 74.5 & 0.31 & 60.6 & 72.8 & 0.33 & 35.9 & 70.6 & 0.36 \\
& CAA             & 64.9 & 77.6 & 0.29 & 59.2 & 75.7 & 0.31 & 34.7 & 73.4 & 0.34 \\
& SEA             & 66.8 & 79.1 & 0.27 & 61.3 & 77.8 & 0.29 & 36.5 & 75.1 & 0.32 \\
& SOP-Role        & 67.5 & 80.4 & 0.26 & 62.1 & 78.5 & 0.28 & 36.8 & 76.3 & 0.31 \\
\addlinespace
& \textbf{ATC}  & \textbf{75.8} & \textbf{92.7} & \textbf{0.12} & \textbf{70.2} & \textbf{91.6} & \textbf{0.14} & \textbf{43.6} & \textbf{89.9} & \textbf{0.16} \\
\midrule
\multirow{8}{*}{S3: Het-12}
& Single-Agent    & 55.1 & 65.8 & 0.43 & 49.9 & 63.5 & 0.46 & 27.9 & 61.8 & 0.48 \\
& Prompt-Only     & 59.2 & 67.4 & 0.41 & 53.6 & 65.9 & 0.43 & 30.7 & 64.1 & 0.45 \\
& Struct-Role     & 62.9 & 72.6 & 0.35 & 56.8 & 70.4 & 0.37 & 32.6 & 68.9 & 0.40 \\
& A-MEM           & 65.7 & 72.1 & 0.34 & 60.3 & 70.2 & 0.36 & 35.5 & 68.5 & 0.39 \\
& CAA             & 64.1 & 74.8 & 0.32 & 58.5 & 72.7 & 0.34 & 34.1 & 70.3 & 0.37 \\
& SEA             & 66.3 & 76.6 & 0.30 & 61.0 & 74.9 & 0.32 & 36.2 & 72.4 & 0.35 \\
& SOP-Role        & 66.8 & 78.2 & 0.29 & 61.5 & 76.1 & 0.31 & 36.4 & 74.0 & 0.33 \\
\addlinespace
& \textbf{ATC}  & \textbf{76.4} & \textbf{93.5} & \textbf{0.11} & \textbf{71.1} & \textbf{92.8} & \textbf{0.13} & \textbf{44.2} & \textbf{91.2} & \textbf{0.15} \\
\bottomrule
\end{tabular}
\end{table*}

Table~\ref{tab:main_results} shows that ATC achieves the best performance
across all benchmarks and team settings. In the homogeneous setting
(S1: Hom-10), ATC improves over the strongest baseline by 6.5 points on
EntCollabBench, 6.8 points on MultiAgentBench, and 5.4 points on SWE-bench
Verified. These gains indicate that explicit role correction improves both role
stability and downstream task success.

The advantage widens in heterogeneous settings. In S2: Het-10, ATC leads the
strongest baseline by 8.3, 8.1, and 6.8 points on EntCollabBench,
MultiAgentBench, and SWE-bench Verified respectively. Heterogeneous teams
appear to amplify role drift:
different model families follow instructions, communicate, and recover from
errors differently, making runtime role control more valuable.

The strongest results appear in S3: Het-12, which has the most agents and at
least five model families, and therefore the most opportunity for role overlap
and authority confusion. There ATC improves the three benchmarks by 9.6, 9.6,
and 7.8 points over the best non-ATC baseline, indicating that it scales to
larger cooperative teams rather than depending on a homogeneous backbone.

Role-control metrics explain the source of the improvement. Across all
settings, ATC obtains the highest RC and the lowest RDE. In S3, for example,
ATC reduces RDE on EntCollabBench from 0.29 for SOP-Role to 0.11, and raises
RC from 78.2 to 93.5. Similar trends hold on MultiAgentBench and SWE-bench
Verified. This ties the task gains to reduced role drift rather than to stronger
prompting or additional memory.

\subsection{Discussion}
\label{ssec:discussion}

The baseline comparison isolates why ATC is effective. Prompt-Only and
Struct-Role declare roles but cannot keep produced behavior inside them; A-MEM
preserves state without determining who may act on it; CAA and SEA edit
representations without separating role alignment from cross-role leakage; and
SOP-Role, the strongest baseline, fixes communication order but cannot repair a
role-violating message once generated.

ATC directly targets this missing mechanism. By scoring each output with
role-subspace deviation, cross-role leakage, linguistic-role violation, and
logical-contract violation, ATC detects role drift before it propagates to
future turns. The correction step then projects the behavior toward a
role-preserving target and regenerates the output under the corrected role
constraint. The largest gains in S2 and S3 support the central claim: explicit
role-contract control becomes more valuable as teams grow more heterogeneous
and the opportunity for authority confusion increases over longer
interactions.

\subsection{Implementation Details}
\label{ssec:impl_details}

Role bases are constructed from role descriptions and role-consistent
demonstrations. The same role descriptions are used for all methods to avoid
prompting advantages. Speech-act vectors are obtained with a fixed classifier
over interaction functions, and logical contracts specify required and
prohibited behaviors for each role. The correction weight $\lambda$ and energy
weights $\alpha,\beta,\gamma,\delta$ are tuned on the validation split and
then fixed across all benchmarks. For CAA and SEA, steering directions are
trained from the same role-consistent and role-drift examples used to build
ATC role bases. All methods use the same prompts, tools, retrieval budget,
context budget, and maximum number of turns.
\subsection{Ablation Studies}
\label{ssec:ablations}

All ablations use EntCollabBench under S3: Het-12, the setting where role
boundaries are most likely to collapse.

\begin{table}[t]
\centering
\caption{\textbf{Component ablation on EntCollabBench, S3: Het-12.}
Each row removes one component of the full method.}
\label{tab:ablation_components}
\small
\setlength{\tabcolsep}{4pt}
\renewcommand{\arraystretch}{0.95}
\begin{tabular}{@{}lccc@{}}
\toprule
\textbf{Variant} & \textbf{Primary $\uparrow$} & \textbf{RC $\uparrow$} & \textbf{RDE $\downarrow$} \\
\midrule
Full ATC & \textbf{76.4} & \textbf{93.5} & \textbf{0.11} \\
\quad w/o role subspace term & 68.2 & 80.1 & 0.27 \\
\quad w/o leakage penalty & 70.4 & 84.3 & 0.22 \\
\quad w/o linguistic term & 72.1 & 87.5 & 0.19 \\
\quad w/o logical contract & 71.6 & 86.8 & 0.20 \\
\quad w/o regeneration & 69.7 & 82.9 & 0.24 \\
\bottomrule
\end{tabular}
\end{table}

Table~\ref{tab:ablation_components} shows that every component contributes.
Removing the role-subspace term causes the largest drop, reducing RC from
93.5 to 80.1 and increasing RDE from 0.11 to 0.27, identifying the geometric
role representation as the principal mechanism. Removing the leakage penalty
also hurts substantially, echoing the directional pattern in the pilot
analysis: agents drift toward other roles, not merely away from their own.
The linguistic and logical terms provide complementary control: the former
catches inappropriate interaction functions, while the latter catches
authority and obligation violations. Removing regeneration weakens the system
because drift is detected but not reliably corrected before entering the
shared context.

\begin{table}[t]
\centering
\caption{\textbf{Correction-objective ablation on EntCollabBench, S3: Het-12.}
The full leakage-regularized projection is most stable.}
\label{tab:ablation_objective}
\small
\setlength{\tabcolsep}{4pt}
\renewcommand{\arraystretch}{0.95}
\begin{tabular}{@{}lccc@{}}
\toprule
\textbf{Correction} & \textbf{Primary $\uparrow$} & \textbf{RC $\uparrow$} & \textbf{RDE $\downarrow$} \\
\midrule
No correction & 59.2 & 67.4 & 0.41 \\
Hard role reminder & 62.9 & 72.6 & 0.35 \\
Pure projection $P_i z$ & 71.3 & 86.1 & 0.20 \\
Hard orthogonality to $\mathcal{F}_i$ & 69.8 & 87.0 & 0.18 \\
Full ATC objective & \textbf{76.4} & \textbf{93.5} & \textbf{0.11} \\
\bottomrule
\end{tabular}
\end{table}

Table~\ref{tab:ablation_objective} isolates the correction rule. Pure
projection improves over prompt-level control, but it does not penalize
cross-role leakage. Hard orthogonality performs worse than the full objective
because valid roles may share semantic directions such as reasoning,
clarification, or evidence use. The full ATC correction is best because it
keeps the behavior inside the assigned role while softly discouraging leakage
into other roles.

\subsection{Stress Tests}
\label{ssec:stress_tests}

We next evaluate robustness under explicit role-drift stressors on
EntCollabBench (S3: Het-12). Each episode receives one deterministic
perturbation. \textbf{Role-confuse} asks an agent
to perform another agent's function. \textbf{Authority-swap} routes a correct
intermediate result through an agent that lacks authority to act on it.
\textbf{Late-correction} revises an earlier decision and tests whether agents
respect the updated division of labor.

\begin{table}[t]
\centering
\caption{\textbf{Role-drift stress tests on EntCollabBench, S3: Het-12.}
Values are task success under perturbation.}
\label{tab:stress_tests}
\small
\setlength{\tabcolsep}{2.5pt}
\renewcommand{\arraystretch}{0.95}
\newcommand{\mhead}[2]{\begin{tabular}[b]{@{}c@{}}\bfseries #1\\\bfseries #2\end{tabular}}
\begin{tabular}{@{}lccc@{}}
\toprule
\textbf{Method} & \mhead{Role-}{confuse $\uparrow$} & \mhead{Authority-}{swap $\uparrow$} & \mhead{Late-}{correction $\uparrow$} \\
\midrule
Prompt-Only & 42.5 & 39.8 & 44.1 \\
Struct-Role & 49.3 & 46.7 & 50.2 \\
A-MEM       & 51.1 & 48.5 & 55.8 \\
SEA         & 54.7 & 51.6 & 56.3 \\
SOP-Role    & 57.9 & 55.2 & 58.6 \\
\textbf{ATC} & \textbf{70.8} & \textbf{68.4} & \textbf{72.1} \\
\bottomrule
\end{tabular}
\end{table}

Table~\ref{tab:stress_tests} shows that ATC is most useful when the system
is actively pushed toward role confusion. SOP-Role is the strongest baseline,
but its fixed communication order cannot repair a role-violating message after
generation. ATC improves success by more than 12 points over SOP-Role on all
three perturbations, with the largest gain on Authority-swap. This supports
the role-contract part of the method: the system must know not only what an
agent can say, but what it is authorized to do.

\subsection{Sensitivity Analysis}
\label{ssec:sensitivity}

We analyze sensitivity to the leakage penalty $\lambda$, which controls how
strongly ATC suppresses movement into other agents' role spaces. Very small
$\lambda$ allows cross-role contamination, while very large $\lambda$ can
over-penalize shared reasoning directions. We evaluate on EntCollabBench (S3: Het-12).

\begin{table}[t]
\centering
\caption{\textbf{Sensitivity to leakage weight $\lambda$ on EntCollabBench,
S3: Het-12.} Moderate leakage control performs best.}
\label{tab:sensitivity_lambda}
\small
\setlength{\tabcolsep}{4pt}
\renewcommand{\arraystretch}{0.95}
\begin{tabular}{@{}lccc@{}}
\toprule
$\boldsymbol{\lambda}$ & \textbf{Primary $\uparrow$} & \textbf{RC $\uparrow$} & \textbf{RDE $\downarrow$} \\
\midrule
0.00 & 71.3 & 86.1 & 0.20 \\
0.25 & 74.2 & 90.4 & 0.15 \\
0.50 & \textbf{76.4} & \textbf{93.5} & \textbf{0.11} \\
1.00 & 75.8 & 92.7 & 0.12 \\
2.00 & 73.6 & 90.8 & 0.15 \\
\bottomrule
\end{tabular}
\end{table}

Table~\ref{tab:sensitivity_lambda} shows that ATC is not overly sensitive
to the exact value of $\lambda$ within a moderate range. The best performance
occurs at $\lambda=0.50$, but $\lambda=0.25$ and $\lambda=1.00$ remain strong.
When $\lambda=0$, the method reduces to pure role projection and leakage
increases. When $\lambda$ is too large, the model becomes overly conservative,
penalizing shared directions that may be useful across roles.

We also test whether ATC remains effective as team size and heterogeneity
increase. We vary the number of agents and model families while keeping the
benchmark, context budget, and task distribution fixed.

\begin{table}[t]
\centering
\caption{\textbf{Sensitivity to team scale and heterogeneity on
EntCollabBench.} ATC remains stable as teams become larger and more diverse.}
\label{tab:sensitivity_scale}
\small
\setlength{\tabcolsep}{4pt}
\renewcommand{\arraystretch}{0.95}
\begin{tabular}{@{}lccc@{}}
\toprule
\textbf{Setting} & \textbf{Primary $\uparrow$} & \textbf{RC $\uparrow$} & \textbf{RDE $\downarrow$} \\
\midrule
Hom-5  & 72.9 & 92.1 & 0.12 \\
Hom-10 & 74.6 & 91.8 & 0.13 \\
Het-10 & 75.8 & 92.7 & 0.12 \\
Het-12 & \textbf{76.4} & \textbf{93.5} & \textbf{0.11} \\
Het-16 & 75.1 & 91.4 & 0.14 \\
\bottomrule
\end{tabular}
\end{table}

Table~\ref{tab:sensitivity_scale} shows that ATC works in both homogeneous
and heterogeneous systems. Performance improves from Hom-5 to Hom-10 because
additional specialization helps task decomposition. Het-10 and Het-12 further
improve because diverse models provide complementary strengths, while ATC
prevents this diversity from turning into role confusion. Het-16 slightly
drops, suggesting that very large teams introduce more communication overhead,
but RC remains above 91\%, showing that role control remains stable.

\subsection{Cost and Latency}
\label{ssec:cost_latency}

ATC adds behavior encoding, drift scoring, role-preserving correction,
constrained regeneration, and re-evaluation before an agent output is released.
We therefore evaluate whether the additional computation is justified by the
resulting role-stability gain. Table~\ref{tab:cost_latency} reports average
model calls, generated tokens, wall-clock latency, and cost-normalized role
consistency improvement on EntCollabBench under the most difficult setting,
S3: Het-12.

We define cost-normalized role gain as
\[
\mathrm{RCG/C}
=
\frac{\mathrm{RC}_{m}-\mathrm{RC}_{\mathrm{Prompt}}}
{\mathrm{Tok}_{m}/10},
\]
where $\mathrm{RC}_{m}$ is the role consistency of method $m$,
$\mathrm{RC}_{\mathrm{Prompt}}$ is the Prompt-Only role consistency, and
$\mathrm{Tok}_{m}$ is the average number of generated tokens per episode in
thousands. Thus, RCG/C measures the role-consistency percentage-point gain over
Prompt-Only per 10K generated tokens.

\begin{table}[t]
\centering
\caption{\textbf{Cost and latency on EntCollabBench, S3: Het-12.}
Tok. = average generated tokens per episode in thousands. Lat. = average
wall-clock latency in seconds. RCG/C = RC gain over Prompt-Only per 10K tokens.}
\label{tab:cost_latency}
\small
\setlength{\tabcolsep}{3.5pt}
\renewcommand{\arraystretch}{0.95}
\begin{tabular}{@{}lcccc@{}}
\toprule
\textbf{Method} & \textbf{Calls $\downarrow$} & \textbf{Tok. $\downarrow$} & \textbf{Lat. $\downarrow$} & \textbf{RCG/C $\uparrow$} \\
\midrule
Prompt-Only & 12.4 & 116.0 & 24.3 & 0.00 \\
Struct-Role & 12.6 & 100.2 & 24.8 & 0.52 \\
A-MEM       & 15.8 & 137.5 & 31.2 & 0.34 \\
CAA         & 12.8 & 119.8 & 27.1 & 0.62 \\
SEA         & 13.1 & 121.4 & 29.3 & 0.76 \\
SOP-Role    & 13.6 & 126.2 & 28.5 & 0.86 \\
ATC       & 15.2 & 132.8 & 32.4 & \textbf{1.97} \\
\bottomrule
\end{tabular}
\end{table}

\begin{figure}[t]
\centering
\includegraphics[width=\columnwidth]{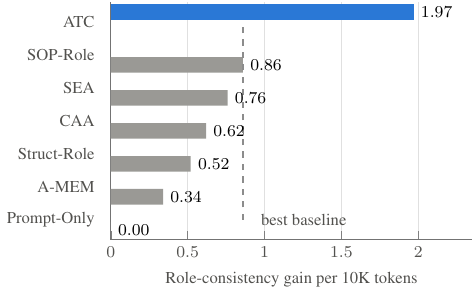}
\caption{\textbf{Cost-normalized role-stability gain.} ATC achieves the
largest role-consistency improvement per 10K tokens, substantially exceeding
the strongest non-ATC baseline.}
\label{fig:cost_normalized_rc}
\end{figure}

Table~\ref{tab:cost_latency} and Figure~\ref{fig:cost_normalized_rc} show
that ATC is more expensive than prompt-only and fixed-workflow baselines, but
its additional computation is efficient relative to the role-stability gain.
ATC uses 132.8K tokens and 15.2 calls per episode, compared with 126.2K
tokens and 13.6 calls for SOP-Role. However, ATC achieves an RCG/C of 1.97,
more than twice the strongest non-ATC baseline. This indicates that the
extra correction and re-evaluation steps are not merely overhead: they provide
the largest role-consistency gain per unit token cost.

\section{Related Work}

ATC's framing draws on three established ideas: design by contract, which specifies obligations and prohibitions for a component~\citep{meyer1992contract}; role-based access control, which ties permissions to roles rather than to individuals~\citep{sandhu1996rbac}; and enforceable security policies, which are exactly those a monitor can decide by observing execution~\citep{schneider2000enforceable}.

Research on LLM agents has developed reasoning--acting loops, tool use, self-reflection, embodied skill acquisition, and multi-agent workflows~\citep{yao2023react,schick2023toolformer,shinn2023reflexion,wang2024voyager,chen2024agentverse,qian2024chatdev}. Multi-agent frameworks and surveys further study orchestration, communication structures, and software-development pipelines~\citep{shen2023hugginggpt,hong2024metagpt,wang2024surveyagents,guo2024multiagentsurvey}. These systems establish the value of specialized agents, but role assignment is usually expressed through prompts, workflow structure, or memory rather than checked as an executable contract on every proposed message. Role-playing benchmarks such as RoleLLM evaluate persona and character fidelity~\citep{wang2024rolellm}, which is distinct from the functional authority ATC enforces, while COMPASS improves long-horizon reasoning through evolving context~\citep{wan2026compass}, a concern orthogonal to authorization.

ATC is also related to representation editing and behavioral steering, including CAA and SEA~\citep{rimsky2024caa,qiu2024sea}. The distinction is the control surface. These methods modify hidden states during generation and therefore require access to the model's internals. ATC instead scores a completed proposal in a functional behavior space, checks linguistic and logical role constraints, and regenerates the message without such access. The question it answers is also different: not whether a generator's internal representation remains coherent, but whether a produced output is permitted for the agent that produced it.

A complementary line of work checks agent behavior rather than steering it. NSF-CoT formally verifies chain-of-thought faithfulness, testing whether each reasoning step is evidentially grounded in the given context and logically valid~\citep{pramanik2026nsfcot}; ATC asks a different question of a different object, namely whether a completed message is authorized for the role that produced it. Closest to our setting, AgentSpec enforces user-specified safety rules over agent actions through a domain-specific language of triggers, predicates, and enforcement mechanisms~\citep{wang2026agentspec}; ATC differs in both object and mechanism, inferring functional speech acts and contract violations from natural-language messages and repairing them by constrained regeneration rather than blocking or substituting a declared action.

\section{Robustness and Deployment Analyses}
\label{sec:additional_analysis}

All methods use identical role prompts, tool access, retrieval and context budgets, maximum turns, and stopping criteria, and CAA and SEA are constructed from the same role-consistent and role-drift examples available to ATC. Under this control, the component ablation in Table~\ref{tab:ablation_components} identifies the geometric role representation as the principal mechanism, with the linguistic and logical terms acting as complementary safeguards rather than carrying the gain on their own.

Appendix~\ref{app:additional_analysis} reports further controls for behavior-encoder capacity, selective correction, legitimate role overlap, and deployment artifacts. The pattern is stable across three BGE v1.5 encoder scales: smaller encoders reduce latency while retaining most of ATC's gain over SOP-Role. Gating regeneration by the pre-correction drift energy produces a smooth reliability--cost trade-off rather than an abrupt collapse. Because role projectors are fixed, they can be precomputed, and the geometric correction is closed form, so most added runtime comes from regeneration and re-evaluation. Finally, ATC does not require mutually orthogonal roles: hard orthogonality and excessive leakage regularization both reduce performance, whereas the validation-selected $\lambda=0.50$ better preserves legitimate shared functions such as reasoning, clarification, and evidence use. These analyses support the interpretation of ATC as a lightweight message-boundary controller whose expensive language-model step can be gated when budgets require it.

\section{Conclusion}

\looseness=-1
ATC addresses a specific coordination failure: a message may be locally plausible yet perform a function that does not belong to the agent that produced it. The method turns role compliance into an explicit runtime contract, combining role-subspace deviation, cross-role leakage, speech-act checks, and logical constraints before a proposed message enters the shared trajectory. Across EntCollabBench, MultiAgentBench, and SWE-bench Verified, ATC improves task performance and role consistency while reducing role-drift energy, with the largest gains in larger and heterogeneous teams. Ablations and stress tests show that the effect does not require strict role orthogonality, and gated regeneration provides a practical reliability--cost trade-off. Because the contract is enforced where messages are exchanged rather than inside any one model, the controller requires only message-level access to the generator, although its auxiliary encoder, classifiers, and contract artifacts remain inspectable. More broadly, the results indicate that deciding which agent may perform which function should be treated as a first-class coordination objective, rather than left to prompting or workflow structure.

\clearpage

\section{Limitations}

ATC provides a runtime mechanism for detecting and correcting role drift at the message boundary, but several limitations remain. First, it depends on the quality of the role representations, role contracts, and behavior encoder used to map agent outputs into functional vectors. If the role basis is incomplete or the contract is underspecified, the controller may miss subtle drift or over-correct behavior that is useful and legitimately shared between roles. Second, black-box deployment relies on role-constrained regeneration rather than direct intervention in the generating model. This broader applicability comes with less direct control over how the revised message is produced.

Third, our evaluation focuses on collaborative multi-agent settings with predefined roles such as planner, retriever, verifier, executor, monitor, and summarizer. While these roles cover many common agentic workflows, real-world systems may contain more fluid or overlapping responsibilities, where strict role boundaries are harder to define. Fourth, role consistency is measured using a fixed rubric and automated judging procedure, which may introduce evaluation bias. Although the same judging protocol is applied to all methods, future work should study more diverse evaluation protocols and robustness across different judges.

Fifth, ATC introduces additional computation through behavior encoding, drift scoring, correction, regeneration, and re-evaluation. Our cost and latency results suggest that this overhead is justified by improved role stability, but the trade-off may be less favorable in very low-latency or resource-constrained deployments. Finally, ATC reduces role drift but does not guarantee complete correctness, safety, or alignment of the overall multi-agent system. A role-consistent output can still be factually incorrect, biased, incomplete, or unsafe. Therefore, ATC should be viewed as a role-stability mechanism rather than a complete solution for reliable or safe autonomous agents.

\section{Ethics Statement}

This work studies role drift in multi-agent LLM systems and proposes ATC as a runtime control method for improving role consistency and collaboration reliability. The intended use of ATC is to reduce unintended role violations, authority confusion, and cross-agent functional leakage in systems that rely on specialized agents. By detecting and correcting role-inconsistent outputs before they enter the shared conversation, ATC may help make multi-agent systems more predictable, auditable, and easier to govern.

The work also introduces potential risks. A role-control mechanism could be misused to enforce overly rigid behavior, suppress useful disagreement, or constrain agents in ways that hide uncertainty or alternative viewpoints. Incorrectly specified role contracts may cause the system to reject valid outputs or preserve harmful role definitions. In high-stakes settings, such as healthcare, law, finance, or safety-critical operations, ATC should not be used as the only safeguard. It should be combined with factual verification, human oversight, domain-specific validation, and safety monitoring.

Our experiments use existing benchmarks and generated multi-agent traces. We do not collect new personal data or conduct human-subject experiments. Any released artifacts should remove personally identifying information from logs and should include documentation describing intended use, evaluation settings, and limitations. We cite the creators of the datasets, benchmarks, and baseline methods used in the study, and all artifacts should be used consistently with their licenses and research-use conditions. If AI assistants were used for writing, editing, or coding support, all final claims, experiments, and analyses should be reviewed and verified by the authors.

\section*{Acknowledgments}

The authors acknowledge the NSF-supported Advanced Cyberinfrastructure Coordination Ecosystem: Services \& Support (ACCESS) program for providing GPU computing resources and UFIT Research Computing for providing HiPerGator computational resources and support.

\bibliography{custom}

\clearpage
\appendix
\renewcommand{\thesection}{A\arabic{section}}
\renewcommand{\thesubsection}{A\arabic{section}.\arabic{subsection}}

\section{Theoretical Properties}
\label{sec:theory}

We establish three basic properties of the proposed correction mechanism.
First, the matrices in Eq.~\eqref{eq:role_projection} and
Eq.~\eqref{eq:forbidden_projection} are valid orthogonal projectors onto the
corresponding role subspaces. Second, the closed-form solution in
Eq.~\eqref{eq:closed_form} solves the correction objective in
Eq.~\eqref{eq:correction_objective}. Third, the correction step satisfies the
local monotonicity property stated in Eq.~\eqref{eq:monotonicity}. These
results apply to the vector-space correction stage. They do not claim that
the full post-regeneration energy always decreases, since the linguistic and
logical terms in Eq.~\eqref{eq:role_energy} depend on the realized text
produced by the regeneration operator in Eq.~\eqref{eq:regeneration}.

\subsection{Projection Validity}
\label{sec:projection_validity}

\paragraph{Proposition 1.}
Let $U_i\in\mathbb{R}^{d\times k_i}$ be any role-basis matrix, possibly with
linearly dependent columns. Then
\[
P_i=U_i(U_i^\top U_i)^\dagger U_i^\top
\]
is the orthogonal projector onto $\operatorname{col}(U_i)$.

\paragraph{Proof.}
For any matrix $U_i$, $U_i(U_i^\top U_i)^\dagger U_i^\top$ is the standard
orthogonal projector onto its column space. We verify the two projector properties.

First, $P_i$ is symmetric. Since $U_i^\top U_i$ is symmetric and positive
semidefinite, its Moore--Penrose pseudoinverse is also symmetric. Therefore,
\[
\begin{aligned}
P_i^\top
&=\left(U_i(U_i^\top U_i)^\dagger U_i^\top\right)^\top \\
&=U_i(U_i^\top U_i)^\dagger U_i^\top
=P_i .
\end{aligned}
\]

Second, $P_i$ is idempotent. Let $G_i=U_i^\top U_i$. Then
\[
P_i^2
=
U_iG_i^\dagger U_i^\top U_iG_i^\dagger U_i^\top
=
U_iG_i^\dagger G_iG_i^\dagger U_i^\top .
\]
By the Moore--Penrose identity
$G_i^\dagger G_iG_i^\dagger=G_i^\dagger$, we obtain
\[
P_i^2
=
U_iG_i^\dagger U_i^\top
=
P_i .
\]
A symmetric idempotent matrix is an orthogonal projector. Its range is
$\operatorname{col}(U_i)$, because $P_i x$ is always a linear combination of
columns of $U_i$, and every vector $u\in\operatorname{col}(U_i)$ satisfies
$P_i u=u$. Hence, $P_i$ is the orthogonal projector onto
$\operatorname{col}(U_i)$. The same argument applies to
$Q_i=V_i(V_i^\top V_i)^\dagger V_i^\top$ in
Eq.~\eqref{eq:forbidden_projection}. \qed

\subsection{Closed-Form Correction}
\label{sec:closed_form_proof}

\paragraph{Proposition 2.}
The coefficient vector in Eq.~\eqref{eq:closed_form} solves the correction
objective in Eq.~\eqref{eq:correction_objective}.

\paragraph{Proof.}
The correction objective in Eq.~\eqref{eq:correction_objective} is
\[
\hat{z}_{i,t}
=
\arg\min_{z'\in\mathcal{S}_i}
\left[
\|z'-z_{i,t}\|_2^2
+
\lambda\|Q_i z'\|_2^2
\right].
\]
Since $\mathcal{S}_i=\operatorname{col}(U_i)$, every feasible vector can be
written as $z'=U_i c$. Substituting this into the objective gives
\[
J(c)
=
\|U_i c-z_{i,t}\|_2^2
+
\lambda\|Q_iU_i c\|_2^2 .
\]
Expanding the objective,
\[
\begin{aligned}
J(c)
&=(U_ic-z_{i,t})^\top(U_ic-z_{i,t}) \\
&\quad+\lambda c^\top U_i^\top Q_i^\top Q_iU_i c .
\end{aligned}
\]
Taking the gradient with respect to $c$ gives
\[
\nabla_c J(c)
=
2U_i^\top(U_ic-z_{i,t})
+
2\lambda U_i^\top Q_i^\top Q_iU_i c .
\]
Setting the gradient to zero yields the normal equation
\[
\left(
U_i^\top U_i
+
\lambda U_i^\top Q_i^\top Q_iU_i
\right)c
=
U_i^\top z_{i,t}.
\]
By Proposition~1, $Q_i$ is an orthogonal projector, so it is symmetric and
idempotent. Hence $Q_i^\top Q_i=Q_i$, and the normal equation becomes
\[
\left(
U_i^\top U_i
+
\lambda U_i^\top Q_iU_i
\right)c
=
U_i^\top z_{i,t}.
\]
When the coefficient matrix is invertible, this system has the usual inverse
solution. In general, it may be singular, so the Moore--Penrose pseudoinverse
gives a minimum-norm solution:
\[
\hat{c}_{i,t}
=
\left(
U_i^\top U_i
+
\lambda U_i^\top Q_iU_i
\right)^\dagger
U_i^\top z_{i,t},
\]
which is exactly Eq.~\eqref{eq:closed_form}. The corrected target is then
$\hat{z}_{i,t}=U_i\hat{c}_{i,t}$. \qed

\subsection{Objective Monotonicity}
\label{sec:monotonicity_proof}

\paragraph{Proposition 3.}
The role-preserving correction satisfies the local monotonicity property in
Eq.~\eqref{eq:monotonicity}:
\begin{align*}
&\|\hat{z}_{i,t}-z_{i,t}\|_2^2
+\lambda\|Q_i\hat{z}_{i,t}\|_2^2 \\
&\leq
\|P_i z_{i,t}-z_{i,t}\|_2^2
+\lambda\|Q_iP_i z_{i,t}\|_2^2 .
\end{align*}

\paragraph{Proof.}
Define the correction objective
\[
J(z')
=
\|z'-z_{i,t}\|_2^2
+
\lambda\|Q_i z'\|_2^2,
\qquad
z'\in\mathcal{S}_i .
\]
By Eq.~\eqref{eq:correction_objective}, $\hat{z}_{i,t}$ is a minimizer of
$J(z')$ over $\mathcal{S}_i$. Therefore, for every feasible
$v\in\mathcal{S}_i$,
\[
J(\hat{z}_{i,t})\leq J(v).
\]
By Eq.~\eqref{eq:role_projection}, $P_i$ projects onto
$\mathcal{S}_i=\operatorname{col}(U_i)$. Hence $P_i z_{i,t}\in\mathcal{S}_i$
and is a feasible comparison point. Setting $v=P_i z_{i,t}$ gives
\[
J(\hat{z}_{i,t})
\leq
J(P_i z_{i,t}).
\]
Substituting the definition of $J$ yields
\begin{align*}
&\|\hat{z}_{i,t}-z_{i,t}\|_2^2
+\lambda\|Q_i\hat{z}_{i,t}\|_2^2 \\
&\leq
\|P_i z_{i,t}-z_{i,t}\|_2^2
+\lambda\|Q_iP_i z_{i,t}\|_2^2 ,
\end{align*}
which proves Eq.~\eqref{eq:monotonicity}. \qed

\subsection{Interpretation of the Bound}
\label{sec:bound_interpretation}

The bound in Eq.~\eqref{eq:monotonicity} shows that the corrected target
$\hat{z}_{i,t}$ is never worse than the ordinary projection $P_i z_{i,t}$
under the leakage-regularized correction objective. The first term preserves
closeness to the original behavior vector from Eq.~\eqref{eq:behavior_vector},
while the second term penalizes leakage into the forbidden-role space defined
by Eq.~\eqref{eq:forbidden_projection}. Thus, the correction simultaneously
maintains role alignment and suppresses cross-role contamination.

This guarantee is intentionally local. It applies to the continuous behavior
vector before regeneration. After applying the constrained regeneration
operator in Eq.~\eqref{eq:regeneration}, the realized text
$\hat{y}_{i,t}$ is re-encoded and re-evaluated. Therefore, the full
role-drift energy in Eq.~\eqref{eq:role_energy} should be measured again on
the released output rather than assumed to decrease.

\section{Reproducibility Details}
\label{app:reproducibility}

This appendix provides the implementation and evaluation details required to
reproduce the ATC experiments. All methods use the same role prompts, tool
interfaces, retrieval budget, context budget, maximum turn limit, and stopping
criteria; only the role-control mechanism differs across methods.

\subsection{Model and Team Configuration}
\label{app:model_config}

Table~\ref{tab:repro_models} summarizes the model composition used in the
three team settings. S1 evaluates whether ATC works when all agents share
the same model behavior. S2 and S3 evaluate whether ATC remains stable when
agents come from different model families and therefore differ in instruction
following, verbosity, uncertainty expression, and tool use.

\begin{table}[t]
\centering
\caption{\textbf{Model and team configuration.} S1 is homogeneous; S2 and S3
are heterogeneous. S3 is the hardest setting, with 12 agents and at least five
model families.}
\label{tab:repro_models}
\small
\setlength{\tabcolsep}{2pt}
\renewcommand{\arraystretch}{0.95}
\begin{tabular}{@{}lccl@{}}
\toprule
\textbf{Setting} & \textbf{Agents} & \textbf{Families} & \textbf{Role Assignment} \\
\midrule
S1: Hom-10 & 10 & 1 & same backbone for all roles \\
S2: Het-10 & 10 & 5 & role groups by model family \\
S3: Het-12 & 12 & 5+ & S2 + auditor, specialist \\
\bottomrule
\end{tabular}
\end{table}

For S1, all agents use the same instruction-tuned backbone. For S2 and S3,
planning, retrieval, verification, execution, and monitoring roles are assigned
to distinct model families. The core ten roles are
\textsc{Coordinator}, \textsc{Planner}, \textsc{Retriever},
\textsc{Memory-Writer}, \textsc{Memory-Reader}, \textsc{Verifier},
\textsc{Critic}, \textsc{Executor}, \textsc{Monitor}, and
\textsc{Summarizer}. S3 additionally includes a \textsc{Policy-Auditor} and a
\textsc{Domain-Specialist}. In the released code, we provide a configuration
file listing each agent's exact checkpoint name, provider, version/date, context window, and role assignment.

\subsection{Decoding and Runtime Settings}
\label{app:runtime_settings}

Table~\ref{tab:repro_runtime} lists the shared decoding and runtime settings.
We keep these fixed across ATC and all baselines to avoid confounding role
control with decoding and context settings.

\begin{table}[t]
\centering
\caption{\textbf{Shared runtime settings.} These settings are fixed across all
methods unless otherwise stated.}
\label{tab:repro_runtime}
\small
\setlength{\tabcolsep}{4pt}
\renewcommand{\arraystretch}{0.95}
\begin{tabular}{@{}lc@{}}
\toprule
\textbf{Parameter} & \textbf{Value} \\
\midrule
Temperature & 0.2 \\
Top-$p$ & 0.95 \\
Max output tokens / turn & 512 \\
Max turns / episode & 200 \\
Context budget & 128K tokens \\
Retrieval budget & top-$k=10$ \\
Seeds & 3 \\
Summary interval & every 20 turns \\
Evaluation split & fixed across all methods \\
\bottomrule
\end{tabular}
\end{table}

The behavior encoder $f$ maps each candidate output, local history, and shared
context into a normalized functional behavior vector. The speech-act mapper
$L$ is a fixed classifier over interaction functions such as
\textsc{assert}, \textsc{retrieve}, \textsc{verify}, \textsc{execute},
\textsc{delegate}, \textsc{summarize}, and \textsc{refuse}. The symbolic
mapper $S$ extracts role-relevant predicates used by the logical contract,
including whether the agent executed an unauthorized action, modified state
without authority, answered outside its role, or failed to satisfy a required
role obligation.

\subsection{ATC Hyperparameters}
\label{app:hyperparameters}

Table~\ref{tab:atc_hyperparams} reports the ATC hyperparameters used in
the main experiments. We tune these values on the validation split and keep
them fixed across all benchmarks and team settings. The same role-consistent
and role-drift examples used to build ATC role bases are also used to train
the steering directions for CAA and SEA.

\begin{table}[t]
\centering
\caption{\textbf{ATC hyperparameters.} Values are selected on validation and
fixed across benchmarks.}
\label{tab:atc_hyperparams}
\small
\setlength{\tabcolsep}{6pt}
\renewcommand{\arraystretch}{0.95}
\begin{tabular}{@{}lcl@{}}
\toprule
\textbf{Symbol} & \textbf{Value} & \textbf{Meaning} \\
\midrule
$\alpha$  & 1.00 & out-of-role deviation weight \\
$\beta$   & 1.00 & cross-role leakage weight \\
$\gamma$  & 0.50 & linguistic violation weight \\
$\delta$  & 0.75 & logical violation weight \\
$\lambda$ & 0.50 & leakage penalty in correction \\
$\rho$    & 0.95 & temporal memory factor \\
$k_i$     & 16   & role-basis rank per agent \\
\bottomrule
\end{tabular}
\end{table}

\subsection{Hardware and Release}
\label{app:hardware_release}

All experiments are run with identical hardware and batching settings across
methods. We report wall-clock latency using end-to-end episode time, including
agent calls, retrieval, ATC scoring, correction, regeneration, and
re-evaluation.

\begin{figure*}[t]
\centering
\includegraphics[width=0.82\textwidth]{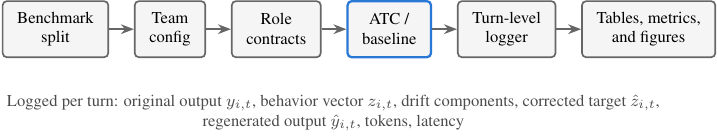}
\caption{\textbf{Reproducibility pipeline.} Each run loads a fixed benchmark
split, team configuration, role contract file, and method configuration. The
logger records agent outputs, behavior vectors, drift scores, corrections,
latency, token usage, and final benchmark metrics.}
\label{fig:repro_pipeline}
\end{figure*}

Figure~\ref{fig:repro_pipeline} summarizes the logging pipeline. Each agent
turn is stored with the original output $y_{i,t}$, behavior vector $z_{i,t}$,
role-drift components, corrected target $\hat{z}_{i,t}$, regenerated output
$\hat{y}_{i,t}$, role-consistency judgment, token usage, and latency. This
allows every aggregate score in the main paper to be recomputed from the
turn-level logs.

\section{Qualitative Examples}
\label{app:qualitative}

Quantitative metrics show that ATC reduces role drift, but qualitative
examples clarify what the correction changes at the turn level. Table~\ref{tab:qual_examples}
shows representative cases from EntCollabBench under S3: Het-12. Each example
contains a drifted output, the ATC-corrected output, and the reason for the
correction.

\begin{table*}[t]
\centering
\caption{\textbf{Qualitative role-drift corrections.} ATC rewrites outputs
so that each agent remains within its assigned function before the message is
released to the shared conversation.}
\label{tab:qual_examples}
\small
\setlength{\tabcolsep}{4pt}
\renewcommand{\arraystretch}{0.98}
\begin{tabular}{@{}p{0.12\textwidth}p{0.27\textwidth}p{0.31\textwidth}p{0.22\textwidth}@{}}
\toprule
\textbf{Agent} & \textbf{Drifted Output} & \textbf{ATC Correction} & \textbf{Why Corrected} \\
\midrule
Verifier &
``We should assign the retrieval task to the Memory-Reader, then ask the
Executor to submit the final answer.'' &
``The proposed plan requires verification. I will check whether the retrieval
step matches the task constraints and whether the Executor has sufficient
evidence before finalization.'' &
Verifier leaked into Planner role. ATC suppresses planning behavior and
restores verification behavior. \\
\addlinespace[4pt]
Retriever &
``The contract likely requires approval from the finance lead, so the answer is
yes.'' &
``I found the relevant policy passage but cannot infer the final decision. The
Verifier should check whether the approval condition is satisfied.'' &
Retriever generated an unsupported conclusion instead of returning evidence. \\
\addlinespace[4pt]
Executor &
``I will update the shared memory to mark this request as approved.'' &
``I can produce the final response only after the Memory-Writer updates the
state and the Verifier confirms the approval condition.'' &
Executor attempted an unauthorized state update. ATC enforces the logical
role contract. \\
\addlinespace[4pt]
Memory-Writer &
``The user should choose option B because it is cheaper.'' &
``I will store the updated constraint: option B has the lowest cost among the
available choices.'' &
Memory-Writer began making a recommendation. ATC restricts it to state
maintenance. \\
\bottomrule
\end{tabular}
\end{table*}

The examples show three common drift patterns. First, an agent may perform a
neighboring role, such as a Verifier becoming a Planner. Second, an agent may
exceed its authority, such as an Executor modifying shared memory. Third, an
agent may transform evidence into a conclusion despite only being authorized to
retrieve or store information. ATC corrects these failures by combining
role-subspace deviation, cross-role leakage, speech-act violation, and
logical-contract violation.

\begin{figure}[t]
\centering
\includegraphics[width=\columnwidth]{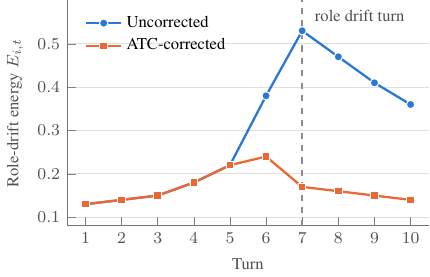}
\caption{\textbf{Example role-drift trace.} The drifted Verifier output causes
a spike in role-drift energy due to cross-role leakage into the Planner
subspace. ATC correction lowers the released output's drift energy before it
enters the shared conversation.}
\label{fig:qual_rde_trace}
\end{figure}

Figure~\ref{fig:qual_rde_trace} illustrates the same phenomenon as a
turn-level trace. The original Verifier output has high cross-role leakage and
therefore high role-drift energy. After correction, the regenerated output
preserves the useful intent of the message but expresses it as a verification
action rather than a planning action. This prevents the local role violation
from becoming shared context for downstream agents.

\begin{figure}[t]
\centering
\includegraphics[width=\columnwidth]{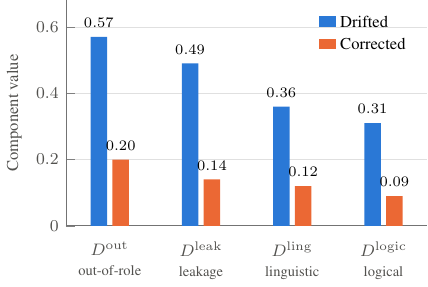}
\caption{\textbf{Component breakdown for a corrected turn.} ATC reduces
out-of-role deviation, cross-role leakage, linguistic violation, and logical
contract violation. The largest reduction occurs in cross-role leakage,
matching the qualitative observation that the Verifier was drifting into the
Planner role.}
\label{fig:qual_component_breakdown}
\end{figure}

Figure~\ref{fig:qual_component_breakdown} breaks down the correction for one
representative turn. The drifted output is fluent and task-relevant, but it
uses the wrong functional mode. The corrected output remains useful while
respecting the assigned role, illustrating why task-level correctness alone is
insufficient for evaluating role-stable multi-agent systems.

\begin{algorithm}[tbp]
\caption{ATC: Agent Trajectory Control}
\label{alg:atc}
\small
\begin{algorithmic}[1]

\Require Agents $\mathcal{A}=\{A_i\}_{i=1}^{n}$ with roles
$\{r_i\}_{i=1}^{n}$
\Require Role bases $U_i\in\mathbb{R}^{d\times k_i}$; encoder $f$;
speech-act mapper $L$; symbolic mapper $S$; regeneration operator $g$
\Require Contracts $C_i=(O_i,B_i)$; linguistic masks $b_i$;
weights $\alpha,\beta,\gamma,\delta$; leakage weight $\lambda$;
memory factor $\rho$
\Ensure Corrected trajectory $\{\hat{y}_{i,t}\}$, corrected drift
scores $\{\hat{E}_{i,t}\}$, and system drift $D_{\mathrm{sys}}(T)$

\Statex \textbf{Precomputation}
\For{each agent $A_i\in\mathcal{A}$}
    \State $P_i\gets U_i(U_i^\top U_i)^\dagger U_i^\top$
    \State $V_i\gets
    \operatorname{basis}\!\left([\,U_j\,]_{j\neq i}\right)$
    \If{$V_i$ has no columns}
        \State $Q_i\gets 0_{d\times d}$
    \Else
        \State $Q_i\gets V_i(V_i^\top V_i)^\dagger V_i^\top$
    \EndIf
    \State $M_i\gets U_i^\top U_i+\lambda U_i^\top Q_iU_i$
    \State $R_i\gets M_i^\dagger U_i^\top$
    \State $D_i(0)\gets 0$
\EndFor

\Statex \textbf{Online control}
\For{$t=1,\ldots,T$}
    \For{each agent $A_i\in\mathcal{A}$}
        \State $D_i(t)\gets \rho D_i(t-1)$
    \EndFor

    \For{each active agent $A_i$}
        \State Generate $y_{i,t}$ from history $h_{i,t}$ and
        shared context $c_t$
        \State $z_{i,t}\gets f(y_{i,t},h_{i,t},c_t)$

        \State $D^{\mathrm{out}}_{i,t}\gets
        \lVert(I_d-P_i)z_{i,t}\rVert_2^2$
        \State $D^{\mathrm{leak}}_{i,t}\gets
        \lVert Q_i z_{i,t}\rVert_2^2$
        \State $D^{\mathrm{ling}}_{i,t}\gets
        \lVert(\mathbf{1}-b_i)\odot L(y_{i,t})\rVert_1$

        \State $s_{i,t}\gets S(y_{i,t},h_{i,t},c_t)$
        \State $D^{\mathrm{logic}}_{i,t}\gets
        \sum_{\phi\in B_i}\mathbf{1}[s_{i,t}\models\phi]$
        \State $D^{\mathrm{logic}}_{i,t}\gets
        D^{\mathrm{logic}}_{i,t}
        +\sum_{\psi\in O_i}\mathbf{1}[s_{i,t}\not\models\psi]$

        \State $E_{i,t}\gets
        \alpha D^{\mathrm{out}}_{i,t}
        +\beta D^{\mathrm{leak}}_{i,t}$
        \State $E_{i,t}\gets E_{i,t}
        +\gamma D^{\mathrm{ling}}_{i,t}
        +\delta D^{\mathrm{logic}}_{i,t}$

        \State $\hat{c}_{i,t}\gets R_i z_{i,t}$
        \State $\hat{z}_{i,t}\gets U_i\hat{c}_{i,t}$
        \State $\hat{y}_{i,t}\gets
        g(y_{i,t},\hat{z}_{i,t},r_i,C_i,b_i,c_t)$

        \State Recompute $\hat{E}_{i,t}$ from $\hat{y}_{i,t}$
        using the same drift terms
        \State $D_i(t)\gets D_i(t)+\hat{E}_{i,t}$
        \State Release $\hat{y}_{i,t}$ and update the shared context
    \EndFor
\EndFor

\State $D_{\mathrm{sys}}(T)\gets
\sum_{i=1}^{n}D_i(T)$
\State \Return $\{\hat{y}_{i,t}\}$,
$\{\hat{E}_{i,t}\}$, and $D_{\mathrm{sys}}(T)$

\end{algorithmic}
\end{algorithm}

\section{Robustness and Deployment}
\label{app:additional_analysis}

This section provides additional analyses of ATC along four dimensions: dependence on auxiliary components and supervision, sensitivity to the behavior encoder, selective correction for lower-cost deployment, and preservation of legitimate overlap between agent roles.

\subsection{Controls and Component Dependence}

ATC does not introduce a task-specific learned controller, nor does it receive privileged supervision relative to the compared methods. Across all experiments, the role prompts, tool access, retrieval budget, context budget, maximum number of turns, and stopping criteria are held fixed. Moreover, the role-consistent and role-drift examples used to construct the ATC role representations are also used to construct the steering directions for CAA and SEA.

The component ablation in Table~\ref{tab:ablation_components} shows graceful degradation when the auxiliary linguistic and logical checks are removed. Without the linguistic term, ATC achieves 72.1 Primary and 87.5 RC; without the logical-contract term, it achieves 71.6 Primary and 86.8 RC. Both variants remain above SOP-Role at 66.8 Primary and 78.2 RC under the same S3: Het-12 setting. Removing the role-subspace term produces the largest degradation, supporting the geometric role representation as the principal control mechanism, while the linguistic mapper $L$ and symbolic mapper $S$ provide complementary checks on speech acts, authority, and obligations.

The runtime computation is also structured to avoid iterative optimization. Because each role basis is fixed during evaluation, the corresponding projectors $P_i$ and $Q_i$ can be precomputed once per agent. Given a candidate behavior vector, Eq.~\eqref{eq:closed_form} then produces the leakage-regularized correction in closed form. As reported in Table~\ref{tab:cost_latency}, relative to SOP-Role, full ATC adds 6.6K generated tokens and 3.9\,s of wall-clock latency per episode (126.2K/28.5\,s versus 132.8K/32.4\,s). This additional computation accompanies an increase in Primary from 66.8 to 76.4 and RC from 78.2 to 93.5, together with a reduction in RDE from 0.29 to 0.11. ATC also obtains the highest cost-normalized role-consistency gain, with RCG/C $=1.97$ compared with $0.86$ for SOP-Role. Constrained regeneration therefore constitutes the primary additional runtime cost, while the geometric correction itself remains inexpensive.

\subsection{Behavior-Encoder Sensitivity}
\label{app:encoder_sensitivity}

We test whether ATC's performance depends critically on the capacity of the fixed behavior encoder. To isolate encoder capacity without introducing a model-family confound, we compare the large, base, and small checkpoints of the BGE v1.5 family
(exact model identifiers in Table~\ref{tab:encoder_sensitivity}).
All experiments use EntCollabBench under S3: Het-12, with the interaction traces, role descriptions, role-consistent demonstrations, logical contracts, basis rank, ATC weights, and evaluation episodes held fixed. Because the checkpoints have different embedding dimensions, the role bases are reconstructed independently in each encoder's native representation space.

\begin{table*}[t]
\centering
\small
\caption{\textbf{Sensitivity to behavior-encoder scale} on EntCollabBench, S3: Het-12. Relative encoding latency is normalized to \texttt{bge-large-en-v1.5}.}
\label{tab:encoder_sensitivity}
\begin{tabular}{lcccc}
\toprule
\textbf{Behavior encoder} & \textbf{Primary $\uparrow$} & \textbf{RC $\uparrow$} & \textbf{RDE $\downarrow$} & \textbf{Relative encoding latency $\downarrow$} \\
\midrule
\texttt{BAAI/bge-large-en-v1.5} & \textbf{76.4} & \textbf{93.5} & \textbf{0.11} & 1.00$\times$ \\
\texttt{BAAI/bge-base-en-v1.5}  & 75.2 & 91.8 & 0.14 & 0.62$\times$ \\
\texttt{BAAI/bge-small-en-v1.5} & 73.6 & 89.4 & 0.17 & \textbf{0.34}$\times$ \\
\bottomrule
\end{tabular}
\end{table*}

Performance degrades smoothly rather than collapsing as encoder capacity is reduced. The small encoder reduces encoding latency by approximately two thirds while retaining 73.6 Primary and 89.4 RC, and the base encoder recovers most of the large-encoder performance at $0.62\times$ relative encoding latency. All three configurations remain above SOP-Role (66.8 Primary, 78.2 RC, and 0.29 RDE). These results indicate that ATC's benefit is not specific to the largest encoder; the essential requirement is an embedding space that preserves distinguishable functional role directions.

\subsection{Gated Correction}
\label{app:gated_atc}

Full ATC corrects every candidate output before it enters the shared
conversation. This design maximizes role stability in extended collaboration,
where a single role-violating output can become context for many subsequent
decisions. For deployments with tighter latency or token budgets, we
additionally evaluate a gated variant in which constrained regeneration is
performed only when the pre-correction drift energy satisfies
$E_{i,t}>\tau$.

We evaluate $\tau\in\{0.20,0.35,0.50\}$ on EntCollabBench under S3: Het-12.
As reported in Table~\ref{tab:atc_motivation}, stable and drifted turns have
mean role-drift energies of $0.36$ and $1.47$, respectively. The selected
thresholds range from below to slightly above the stable-turn mean while
remaining below the drifted-turn mean, allowing us to vary how selectively
ATC invokes constrained regeneration.

\begin{table*}[t]
\centering
\small
\caption{\textbf{Gated ATC cost--performance trade-off} on EntCollabBench,
S3: Het-12. Full ATC corrects every turn, whereas ATC-Gated regenerates only
when $E_{i,t}>\tau$.}
\label{tab:gated_atc}
\setlength{\tabcolsep}{4pt}
\begin{tabular}{lrrrrrr}
\toprule
\textbf{Correction policy} &
\textbf{Corrected turns (\%)} &
\textbf{Primary $\uparrow$} &
\textbf{RC $\uparrow$} &
\textbf{RDE $\downarrow$} &
\textbf{Tokens (K) $\downarrow$} &
\textbf{Latency (s) $\downarrow$} \\
\midrule
SOP-Role (no runtime correction)
& 0.0 & 66.8 & 78.2 & 0.29 & \textbf{126.2} & \textbf{28.5} \\
ATC-Gated, $E_{i,t}>0.50$
& 24.7 & 71.3 & 85.1 & 0.21 & 128.1 & 29.7 \\
ATC-Gated, $E_{i,t}>0.35$
& 46.8 & 73.9 & 89.2 & 0.16 & 129.9 & 30.8 \\
ATC-Gated, $E_{i,t}>0.20$
& 72.6 & 75.6 & 92.1 & 0.13 & 131.5 & 31.7 \\
\addlinespace
ATC (correction at every turn)
& 100.0 & \textbf{76.4} & \textbf{93.5} & \textbf{0.11}
& 132.8 & 32.4 \\
\bottomrule
\end{tabular}
\end{table*}

Table~\ref{tab:gated_atc} shows a smooth cost--performance trade-off.
Increasing the threshold reduces the fraction of regenerated turns, generated
tokens, and latency, while progressively lowering task performance and role
consistency. At $\tau=0.20$, ATC corrects 72.6\% of turns, reducing usage to
131.5K tokens and 31.7\,s while retaining 75.6 Primary, 92.1 RC, and 0.13 RDE.
Higher thresholds provide further cost reductions but allow additional role
drift. The all-turn ATC configuration remains strongest in Primary, RC, and
RDE, whereas the gated variants provide lower-cost operating points when
exhaustive correction is unnecessary.

\subsection{Role Overlap and Over-Correction}
\label{app:overcorrection}

ATC does not assume mutually exclusive or orthogonal role spaces. Different roles may legitimately share functional directions corresponding to reasoning, evidence use, clarification, or uncertainty expression. The leakage term in Eq.~\eqref{eq:correction_objective} is therefore a soft penalty rather than a hard orthogonality constraint. The correction-objective ablation in Table~\ref{tab:ablation_objective} supports this choice: enforcing hard orthogonality to the forbidden-role space yields 69.8 Primary, compared with 76.4 for the full leakage-regularized objective. Similarly, Table~\ref{tab:sensitivity_lambda} shows that excessive leakage suppression is detrimental: increasing $\lambda$ from the validation-selected $0.50$ to $2.00$ decreases Primary from 76.4 to 73.6 and increases RDE from 0.11 to 0.15.

To examine whether these changes correspond to harmful suppression of legitimate shared behavior, we conduct a targeted audit of 100 held-out turns containing valid overlap between agent responsibilities. We compare the selected $\lambda=0.50$ with the deliberately over-constrained $\lambda=2.00$.

\begin{table*}[t]
\centering
\small
\caption{\textbf{Audit of legitimate role overlap} on held-out EntCollabBench turns. Moderate leakage control substantially reduces harmful over-correction while preserving task-relevant content.}
\label{tab:overcorrection}
\begin{tabular}{lccccc}
\toprule
\textbf{Leakage weight} &
\textbf{Primary $\uparrow$} &
\textbf{RC $\uparrow$} &
\textbf{RDE $\downarrow$} &
\textbf{Harmful over-correction $\downarrow$} &
\textbf{Content preserved $\uparrow$} \\
\midrule
$\lambda=0.50$ & \textbf{76.4} & \textbf{93.5} & \textbf{0.11} & \textbf{5.6\%} & \textbf{94.3\%} \\
$\lambda=2.00$ & 73.6 & 90.8 & 0.15 & 18.4\% & 82.1\% \\
\bottomrule
\end{tabular}
\end{table*}

At $\lambda=0.50$, harmful over-correction occurs in only 5.6\% of the audited cases and 94.3\% of task-relevant content is preserved. Under the intentionally stronger $\lambda=2.00$ penalty, harmful corrections rise to 18.4\% and content preservation falls to 82.1\%. This behavior is consistent with the accompanying decrease in Primary and increase in RDE. Together with the hard-orthogonality ablation, the audit indicates that validation-selected soft leakage control suppresses unauthorized role transfer without requiring semantic isolation between agents.

\subsection{Per-Domain Role Artifacts and Curation}
\label{app:role_artifacts}

Instantiating ATC in a new domain requires three groups of per-role artifacts: (i) a role description, (ii) role-consistent demonstrations, and (iii) linguistic and logical constraints specifying permitted interaction functions together with required and prohibited behaviors. No task-specific controller is trained. Role bases are constructed from the role descriptions and role-consistent demonstrations; once a basis $U_i$ is available, its role projector $P_i$ is computed automatically using Eq.~\eqref{eq:role_projection}, and the corresponding forbidden-role projector $Q_i$ is precomputed from the other role bases.

The demonstrations are not an ATC-specific supervision advantage: the same role-consistent and role-drift examples are provided to CAA and SEA for constructing their steering directions. The released artifact comprises the role prompts, role contracts, role-basis construction scripts, benchmark wrappers, evaluation scripts, hyperparameter files, and anonymized turn-level logs. Consequently, adapting ATC to a new domain primarily requires specifying the role artifacts rather than training a new controller. Per-role demonstration counts, logical-predicate counts, and curation time should be reported alongside these configurations to make this authoring cost explicit.

\end{document}